\documentclass{article}
\usepackage{ijcai24}

\usepackage{times}
\usepackage{soul}
\usepackage{url}
\usepackage[hidelinks]{hyperref}
\usepackage[utf8]{inputenc}
\usepackage[small]{caption}
\usepackage{graphicx}
\usepackage{amsmath}
\usepackage{amsthm}
\usepackage{booktabs}
\usepackage{algorithm}
\usepackage{algorithmic}
\usepackage[switch]{lineno}

\usepackage{natbib}
\usepackage{amssymb}
\usepackage{mathtools}
\usepackage{tabularx}
\usepackage{multirow}
\usepackage{multicol}
\usepackage{booktabs}
\usepackage{subcaption}
\newtheorem{definition}{Definition}

\usepackage{lipsum}
\usepackage{xcolor}
\usepackage{enumitem}
\usepackage{comment}
\usepackage{bbding}

\usepackage{cancel}

\DeclareMathOperator{\fcomp}{\bigcirc}
\newcommand{\Hquad}{\hspace{0.2em}} 

\newcounter{eqfn}
\def\equalcontrib{%
      \ifnum\value{eqfn}=0%
        \footnote{These authors contributed equally.}%
        \setcounter{eqfn}{\value{footnote}}%
      \else%
        \footnotemark[\value{eqfn}]%
      \fi%
    }%

\title{StableKD: Breaking Inter-block Optimization Entanglement\\ for Stable Knowledge Distillation}

\author{
Shiu-hong Kao\equalcontrib
\and
Jierun Chen\equalcontrib
\and
S.H. Gary Chan\\
\affiliations
The Hong Kong University of Science and Technology\\
\emails
\{skao, jcheneh, gchan\}@cse.ust.hk
}

\begin{document}

\makeatletter
\let\@oldmaketitle\@maketitle
\renewcommand{\@maketitle}{\@oldmaketitle
  \vspace{-0.2in}
  \includegraphics[width=\linewidth]{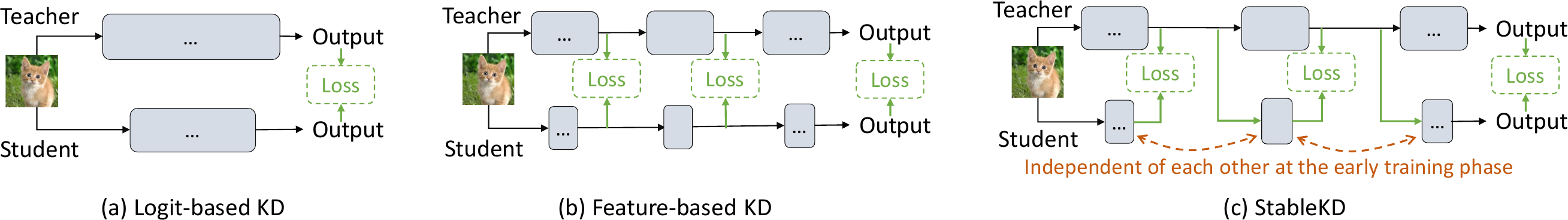} 
  \captionof{figure}{Comparison of StableKD with vanilla logit-based and feature-based knowledge distillation frameworks.}\label{fig:comparison}\bigskip}

\makeatother

\maketitle

\begin{abstract}
Knowledge distillation (KD) has been recognized as an effective tool to compress and accelerate models. However, current KD approaches generally suffer from an accuracy drop and/or an excruciatingly long distillation process. In this paper, we tackle the issue by first providing a new insight into a phenomenon that we call the \emph{inter-block optimization entanglement} (IBOE), which makes the conventional end-to-end KD approaches unstable with noisy gradients. We then propose StableKD, a novel KD framework that breaks the IBOE and achieves more stable optimization. StableKD distinguishes itself through two operations: \emph{Decomposition} and \emph{Recomposition}, where the former divides a pair of teacher and student networks into several blocks for separate distillation, and the latter progressively merges them back, evolving towards end-to-end distillation. We conduct extensive experiments on CIFAR100, Imagewoof, and ImageNet datasets with various teacher-student pairs. Compared to other KD approaches, our simple yet effective StableKD greatly boosts the model accuracy by 1\% -- 18\%, speeds up the convergence up to $10\times$, and outperforms them with only $40\%$ of the training data.
\end{abstract}
\section{Introduction}
While the great potential of deep learning has been showcased by many large-scale models running on the cloud~\citep{devlin2018bert,holcomb2018overview}, a burgeoning direction is to bring the intelligence to the edge, by building compact and computation-efficient models. Knowledge Distillation (KD), among various branches
of methods, is well-known for its conceptual simplicity and ease of implementation. KD aims to distill knowledge from a large-scale teacher model to a smaller and faster student model. A typical vanilla KD approach provides the informative ``soft targets'' from the last layer of the teacher to the student~\citep{hinton2015distilling}. The feature-based KD approaches further enforce the alignment of intermediate feature maps~\citep{heo2019comprehensive,jung2021fair}. Though effective in downsizing the model and reducing the computational complexity, they generally lead to deterioration in accuracy, leaving an accuracy gap between the teacher and the student. \cite{beyer2022knowledge} narrow down the gap by distilling patiently for a large number of epochs. However, such an excruciatingly long distillation process is costly in terms of computational resources and energy. The situation can worsen when the task becomes more complex, involving a huge training dataset. Collectively, these factors diminish the appeal of KD for intensive and practical deployment.

\emph{How can we improve a student's accuracy and simultaneously accelerate the KD process?} 
We probe into the problem from a closely related optimization perspective. Given a student network composed of a series of blocks\footnotemark[1], current KD approaches mainly train it in an end-to-end manner. Therefore, the optimization of a block is conditioned on or entangled with other blocks by noisy gradients. We call this phenomenon \emph{inter-block optimization entanglement} (IBOE). We hypothesize that such entanglement can incur optimization instability especially when the randomly initialized parameters are far from convergence at the early training phase. The optimization instability will then lead to accuracy fluctuation and slow down the convergence~\citep{shalev2017failures}.\footnotetext[1]{A block refers to a layer or a stack of consecutive layers in a neural network.}

The above observation motivates us to propose a novel KD framework that has a more stable optimization -- StableKD. It consists of two operations, \emph{Decomposition} and \emph{Recomposition}. \emph{Decomposition} breaks the IBOE by dividing a pair of teacher and student networks into separate blocks. Notably the input of a student block is identical to the input of its corresponding teacher block, rather than the output of its preceding student block. This simple design essentially differentiates StableKD from prior works, as shown in Figure~\ref{fig:comparison}. It enables the student blocks to learn independently, stably, and rapidly from their teacher. However, it incurs an architectural discrepancy between training and inference, i.e., the student blocks are blind to each other during training but have to collaborate for inference. Such discrepancy, as verified in the paper, can adversely affect the student's final accuracy. To overcome this side effect, \emph{Recomposition} is imposed to progressively merge the divided blocks back along the training process. This is reasonable as the student blocks get more mature after certain epochs, and the IBOE becomes less of a concern. 

As such, we keep our StableKD framework simple yet effective in increasing the optimization stability, and thereby boosting the model accuracy and accelerating the KD process. We conduct experiments on the widely adopted CIFAR100~\citep{krizhevsky2009learning}, Imagewoof~\citep{imagewang}, and ImageNet~\citep{russakovsky2015imagenet} datasets, using diverse combination of teacher and student pairs. The results show that StableKD consistently outperforms the state of the art by 1\% -- 18\% in absolute accuracy, especially when constrained by a limited amount of training budget. Equipped with StableKD, the Swin-T network~\citep{liu2021swin} achieves impressive 82.6\% accuracy on ImageNet, much better than its 81.2\% counterpart with vanilla KD. When having similar accuracy, StableKD can also speed up the KD process up to 10$\times$. For example, StableKD reaches 78.45\% accuracy on CIFAR100 with 20 epochs while vanilla KD requires 200 epochs to reach 78.07\%. Moreover, our StableKD framework enjoys further benefits of being data-efficient. For example, with only 40\% of the CIFAR100 or 60\% of the ImageNet training data, StableKD attains comparable or even higher accuracy than vanilla KD with the whole dataset.

To summarize, our contributions are as follows:
\begin{itemize}
\itemsep0em 
\item We reveal and verify, for the first time, the IBOE issue commonly occurred in end-to-end KD approaches.
\item We present StableKD, a simple yet effective KD framework, to mitigate the IBOE issue.
\item We conduct extensive experiments and verify that our StableKD framework is highly accurate, fast, and data-efficient. It is a generic framework and can be applied to various tasks and neural networks.
\end{itemize}
\section{Related Work}
Knowledge distillation (KD), introduced by \cite{hinton2015distilling}, is a prominent technique for model compression that leverages a large teacher network to guide a smaller student network~\citep{hinton2015distilling, li2017mimicking, li2021online}. Existing works can be broadly classified into two types based on the knowledge transferred: logit-based and feature-based. The former imposes various regularization on the student’s output~\citep{hinton2015distilling,zhao2022decoupled,furlanello2018born,mirzadeh2020improved}, while the latter further transfers intermediate representations~\citep{zagoruyko2016paying,heo2019knowledge,huang2017like,kim2018paraphrasing,shen2019amalgamating,guan2020differentiable,chen2021cross,romero2014fitnets,peng2019correlation,chen2021distilling}. 
  
Though promising in transferring diverse knowledge from a teacher to its student, 
those KD approaches requires intensive computation and time cost to maintain the teacher's performance~\citep{beyer2022knowledge}. 
 Some recent works have attempted to accelerate the KD process~\citep{yun2021re,shen2022fast}, for instance, by caching the teacher’s output for later use in the KD process~\citep{shen2022fast}. However, these approaches incur storage overhead, performance degradation, and limited acceleration rate (usually less than 2$\times$). By contrast, our StableKD overcomes these drawbacks and achieves a more substantial improvement in reducing the distillation time.

Another line of research is to enhance the data efficiency of KD, 
by either updating a subset of parameters alternately~\citep{wang2018progressive,kulkarni2019data}, adding cross data flow as regularization~\citep{bai2020few,shen2021progressive}, or expanding the training set~\citep{wang2020neural,nguyen2022black,laskar2020data}. Similar to those works, our StableKD benefits from high data efficiency by enabling blockwise KD. However, our StableKD is much simpler to implement, without complex data or architectural manipulations. Our StableKD is also faster to train, as it performs the blockwise KD almost parallelly.

\section{Hypothesis and Experimental Validation}\label{hypothesis}
\begin{figure*}[t!]
    \centering
    \begin{subfigure}[b]{0.31 \linewidth}
        \centering
        \includegraphics[width=\linewidth]{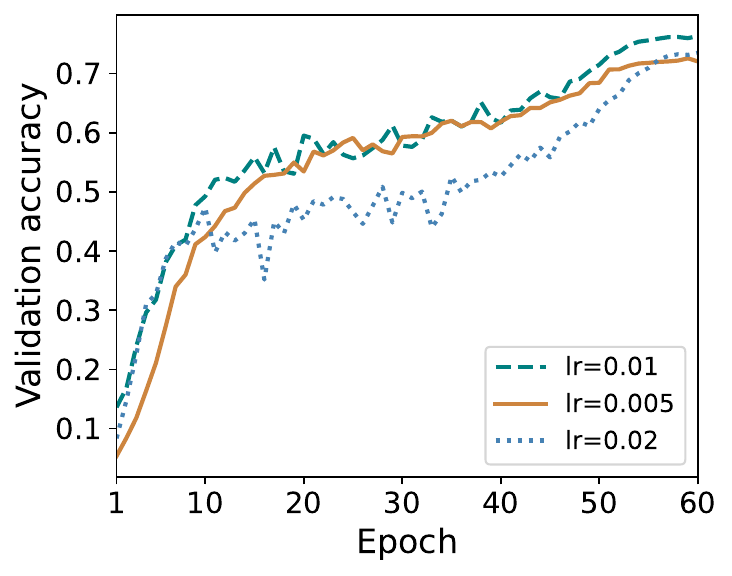}
        \caption{Validation accuracy increases with fluctuations for most of the epochs and for all runs with different learning rates.}
        \label{fig:hypothesis_a}
    \end{subfigure}
    \hfill
    \begin{subfigure}[b]{0.33 \linewidth}
        \centering
        \includegraphics[width=\linewidth]{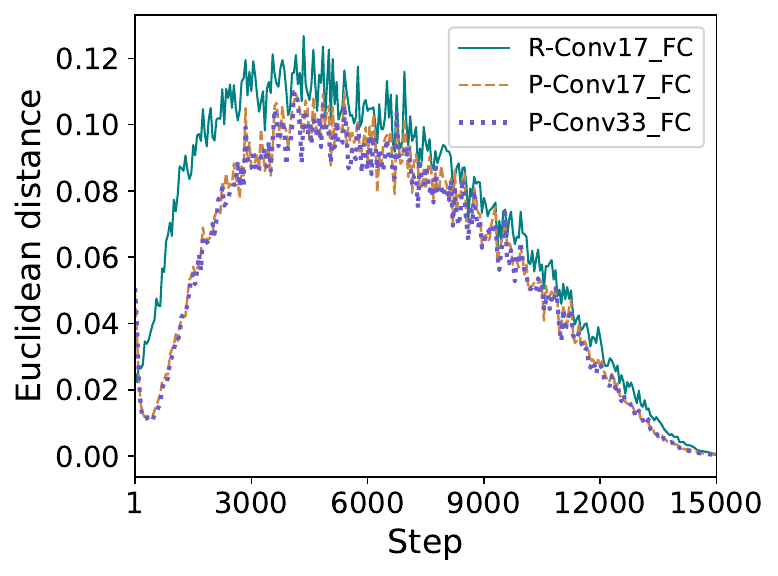}
        \caption{The randomly initialized R-Conv17 incurs higher Euclidean distance of the FC parameters, than the pre-trained counterparts.}
        \label{fig:hypothesis_b}
    \end{subfigure}
    \hfill  
    \begin{subfigure}[b]{0.335 \linewidth}
        \centering
        \includegraphics[width=\linewidth]{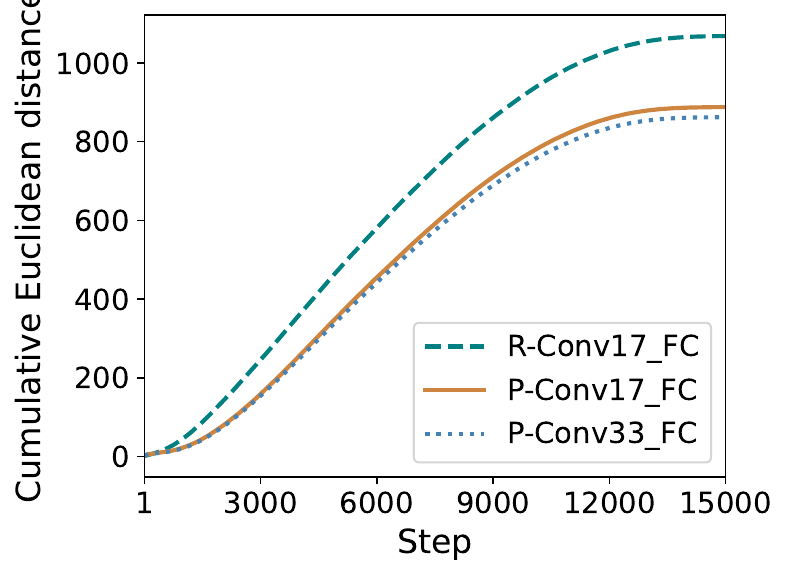}
        \caption{Further supporting the conclusion drawn in (\subref{fig:hypothesis_b}), yet from the angle of cumulative Euclidean distance.}
        \label{fig:hypothesis_c}
    \end{subfigure}
    \caption{Results of the preliminary experiments. All point to the soundness of our IBOE hypothesis.}
    \label{fig:hypothesis}
\end{figure*}
It is commonly believed that the success of deep learning largely lies in the nature of being deep, \emph{i.e.}, the more layers stacked, the better performance a neural network could achieve. However, optimizing very deep neural networks is challenging for various reasons, such as the problems of gradient explosion~\citep{iskoldskii1977instability}  and gradient vanishing~\citep{hochreiter1998vanishing}. Specifically, \cite{shalev2017failures} identify the issue of noisy gradient in end-to-end training, which slows down the convergence significantly. Inspired by this, we investigate the cause of 
noisy or non-informative gradients and propose the following hypothesis:

\paragraph{The IBOE Hypothesis.} 
\textit{For each block in a randomly-initialized neural network, the randomness of its back-propagated gradients is influenced by the random noises in the parameters of its adjacent blocks. This causes optimization instability, reduces the network accuracy, and delays the convergence, especially when the network is far from mature. We refer to this phenomenon as \textbf{I}nter-\textbf{b}lock \textbf{O}ptimization \textbf{E}ntanglement (IBOE).}

To empirically show that IBOE exists, we conduct two simple exploratory experiments: 1) Monitor a student network's validation accuracy during the vanilla KD process, and check if there's any fluctuations occurred, as a proxy to qualitatively reflect the optimization stability. 2) Monitor the Euclidean distance between the parameters of two consecutive steps, as another proxy to quantitatively measure the optimization stability~\citep{hardt2016train}.

For the first toy experiment, we use the widely-adopted CIFAR100~\citep{krizhevsky2009learning} dataset, and use the golden ResNet34 and ResNet18~\citep{he2016deep} as the teacher and student networks, respectively. We then conduct the vanilla KD (refer to Section~\ref{subsec:vanilla_kd} for more details) for 60 epochs using the SGD optimizer~\citep{pytorch} and the OneCycle scheduler~\citep{smith2019super}. The base learning rate is set as 0.005, 0.01, and 0.02 for three different runs. 
From Figure~\ref{fig:hypothesis}(\subref{fig:hypothesis_a}), we can clearly see that their validation accuracy increases with fluctuations for most of the epochs. One may argue that it is because the step size or the learning rate is too large and overshoots in the optimal direction. However, such fluctuations still exist for relatively smaller learning rate. Indeed, this phenomenon can be interpreted as  
the consequence of the IBOE issue, where different blocks collaborate but also interfere with each other. 

To further validate our hypothesis quantitatively, we conduct the second experiment using three feature extractors: R-Conv17, P-Conv17, and P-Conv33. The prefixes ``R-'' and ``P-'' indicate whether a network is randomly initialized or pre-trained, while ``Conv17'' and ``Conv33'' denote the convolutional layers of a ResNet18 and a ResNet34, respectively, both excluding the original fully connected (FC) layer. We then attach a new FC layer as the classifier to each of them. Note that the parameters of the three FC layers have the same initial random values. We train them following the similar setting to our first experiment, but now we monitor the Euclidean distance between the FC parameters of two consecutive steps, as well as its cumulative version.
As shown in Figure~\ref{fig:hypothesis}~(\subref{fig:hypothesis_b}) and  (\subref{fig:hypothesis_c}), we highlight three key points: 
\begin{enumerate}[label={\arabic*)}]
    \item With R-Conv17, the classifier FC suffers from larger distances and thereby higher instability than those with P-Conv17 and P-Conv33.
    \item This gap is much larger in the early training steps.
    \item P-Conv33 has a similar distance level to P-Conv17, showing that the optimization stability is less relevant to the network capacity but more restricted by the mature state or noise level of the parameters.
\end{enumerate}
In short, all these results are consistent to our IBOE hypothesis, and motivates our next step to address the issue.

\section{StableKD: Stable Knowledge Distillation}
This section begins with a brief overview of the vanilla KD method, which is a standard baseline in the field. Next, we present our StableKD framework in detail, focusing on the loss functions and the algorithm design.

\subsection{Preliminary: Vanilla KD} \label{subsec:vanilla_kd}
The vanilla KD approach proposed by \cite{hinton2015distilling} is well-known for its simplicity and effectiveness. One typical task is the image classification, where a student network learns to imitate a teacher by predicting soft targets, \emph{e.g.}, assigning an image a 0.8 probability of being a cat and a 0.2 probability of being a grassland. We present the mathematical formulation of this process below.

An input image $x$ is fed into a teacher network $f^\mathcal{T}(\cdot|\theta^\mathcal{T})$ with parameters $\theta^\mathcal{T}$ as well as a student network $f^\mathcal{S}(\cdot|\theta^\mathcal{S})$ with parameters $\theta^\mathcal{S}$. Their outputs are denoted as $\hat{y}^\mathcal{T}= \text{Softmax}\left(f^\mathcal{T}(x|\theta^\mathcal{T})\right)$
and $\hat{y}^\mathcal{S}=\text{Softmax}\left(f^\mathcal{S}(x|\theta^\mathcal{S})\right)$ correspondingly, which are then compared with the ground truth label $y$ to compute the training loss. Note that the notation of Softmax temperature is ignored for simplicity, where value $1$ is used for all experiments in this paper. There are two loss terms involved in the vanilla KD, the cross-entropy (CE) loss and the Kullback–Leibler (KL) divergence. Both serve as a measure of statistical distance, but the  former is calculated between $\hat{y}^\mathcal{S}$ and $y$ while the latter is between $\hat{y}^\mathcal{S}$ and $\hat{y}^\mathcal{T}$, \emph{i.e.,}
 \begin{equation} \label{eq:KL_loss} \ell_{CE}=\mathcal{L}_{CE}\left(\hat{y}^\mathcal{S}, y\right),
\end{equation}
 \begin{equation} \label{eq:CE_loss} \ell_{KL}=\mathcal{L}_{KL}\left( \hat{y}^\mathcal{S}, \hat{y}^\mathcal{T}\right).
\end{equation}
Now the overall vanilla KD loss can be calculated by
\begin{equation} \label{eq:KD}
\mathcal{L}_{KD}=(1-\alpha)\ell_{CE}+\alpha\ell_{KL},
\end{equation}
where $\alpha \in (0,1)$ is a hyper-parameter balancing the above two loss terms.

\subsection{Details of StableKD}
\begin{figure}[t]
    \centering
    \includegraphics[width=\linewidth]{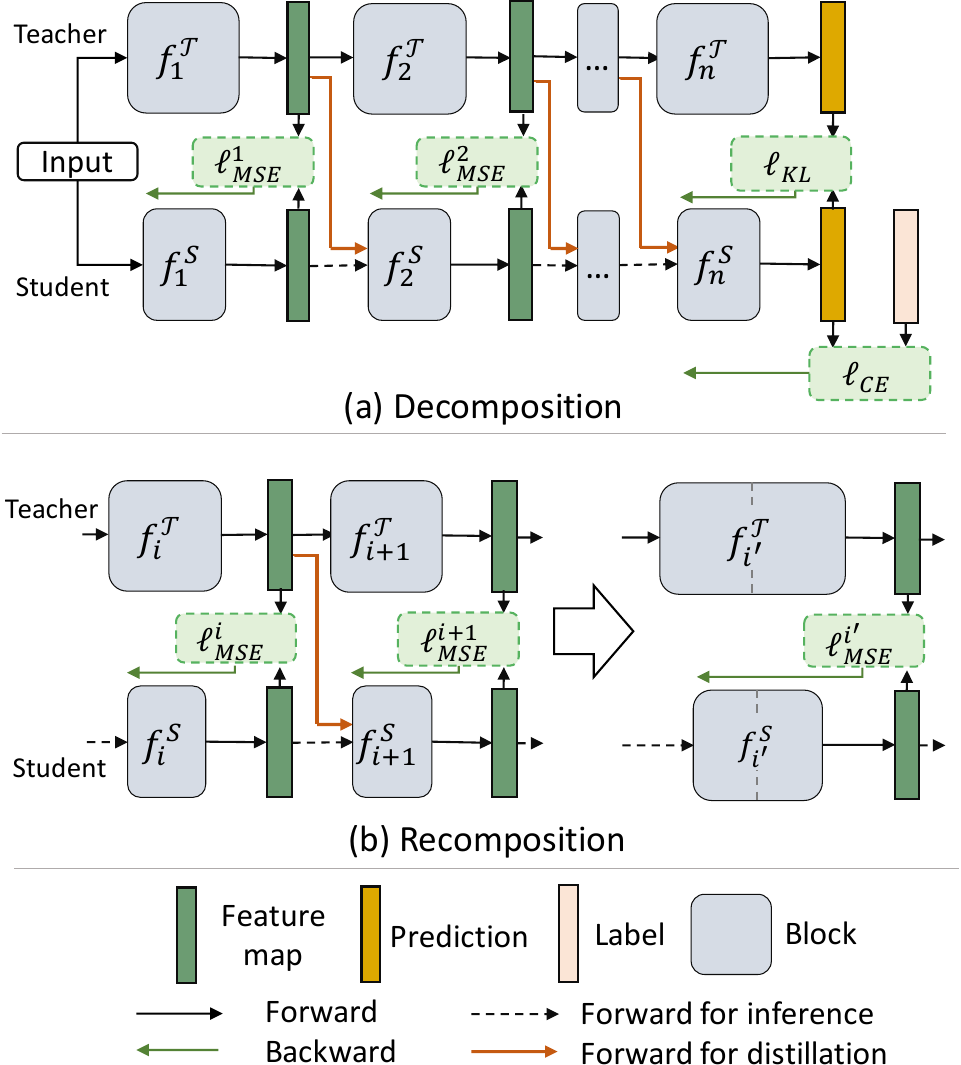}
    \caption{Overview of our StableKD. It has two key operations, \emph{Decomposition} and \emph{Recomposition}. The former divides the teacher and student networks into several pairs of blocks, which then perform the KD process separately for certain epochs. After that, the latter merges the adjacent blocks two by two, followed by a new KD stage.}
    \label{fig:design}
\end{figure}
The above formulation shows that the vanilla KD naively treats the student network as a whole and a black box for end-to-end training, which inherently suffers from the IBOE issue as we have examined in Section~\ref{hypothesis}. Building upon this finding, we introduce StableKD, a simple yet effective KD framework that enables more fine-grained block-wise training to mitigate the IBOE problem. Specifically, our StableKD involves two novel operations \emph{Decomposition}  and \emph{Recomposition}. The \emph{Decomposition} operation, as shown in Figure~\ref{fig:design}~(a), divides a pair of teacher and student networks into several blocks for separate distillation. Note that the intermediate feature maps from the teacher are directed to the subsequent block of the student. By contrast, the \emph{Recomposition} operation, as shown in Figure~\ref{fig:design}~(b), progressively merges the decomposed blocks back along the training process, to preserve the architectural consistency between training and testing, avoiding potential accuracy drop. To elaborate our workflow more clearly, we next provide a rigorous mathematical description.

\subsubsection{Decomposition.}\label{divide}
A neural network $f(\cdot|\theta)$ with parameters $\theta$ can be 
decomposed into multiple
blocks $f_1, f_2, ..., f_k$ satisfying $f(\cdot|\theta) = f_k\circ ...\circ f_2 \circ f_1(\cdot|\theta)$ for some $k\in \mathbb{N}$, where $f\circ g(\cdot)\coloneqq f(g(\cdot))$.  For simplicity, we denote 
$\overset{k}{\underset{i=1}{\fcomp}} f_i \coloneqq f_k \circ f_{k-1} \circ ... \circ f_1.$ As there are different ways of decomposition, we denote $\mathcal{D}$ as one specific decomposition strategy, which specifies the number of divided blocks $k$ and how these blocks are partitioned from the student and teacher models. $\mathcal{D}$ can be empirically determined and preset at the beginning of StableKD process.
In this paper, we consider the FC layer at the end as a single block and divide the remaining layers, \emph{e.g.}, convolutional layers and multi-head attention layers into 2 -- 4 blocks, based on their feature dimensions and receptive fields (refer to appendix for more details).

After the decomposition, we get multiple pairs of teacher and student blocks. Training is then conducted separately for each pair. For the last pair, the training loss is similar to Equation~\ref{eq:KD}, comprising a CE loss and a KL divergence:
\begin{equation} \label{eq:StableKD_CE}
    \ell_{CE}= \mathcal{L}_{CE}(\hat{y}^{\mathcal{T}, \mathcal{S}},y),
\end{equation}
\begin{equation}\label{eq:StableKD_KL}
    \ell_{KL}= \mathcal{L}_{KL}(\hat{y}^{\mathcal{T}, \mathcal{S}},\hat{y}^{\mathcal{T}}),
\end{equation}
where $\hat{y}^{\mathcal{T}, \mathcal{S}}$ denotes the prediction of the last student block, which relies on its teacher's preceding blocks, \emph{i.e.}, $ \hat{y}^{\mathcal{T}, \mathcal{S}}=\text{Softmax}\left(f^\mathcal{S}_k\circ\overset{k-1}{\underset{i=1}{\fcomp}} f^\mathcal{T}_i(x|\theta^\mathcal{S}, \theta^\mathcal{T}, \mathcal{D})\right)$. 
For the remaining pairs of blocks, we adopt the mean squared error (MSE) loss to densely transfer the knowledge. Specifically, the MSE loss $\ell_{MSE}^{\Hquad i}$ for block $f_i$ is written as 
\begin{equation} \label{eq:StableKD_MSE1}
    \ell_{MSE}^{\Hquad 1}= \mathcal{L}_{MSE}\left(f^\mathcal{T}_1(x|\theta^\mathcal{T}, \mathcal{D}), f^\mathcal{S}_1(x|\theta^\mathcal{S}, \mathcal{D})\right),
\end{equation}
for $i=1$, and 
\begin{equation} \label{eq:StableKD_MSEi}
    \begin{aligned}
    \ell_{MSE}^{\Hquad i}= &\mathcal{L}_{MSE}\Bigl(\operatorname*{\fcomp}_{j=1}^{i} f^\mathcal{T}_j(x|\theta^\mathcal{T}, \mathcal{D}),\\
    &f^\mathcal{S}_i\circ \operatorname*{\fcomp}_{j=1}^{i-1} f^\mathcal{T}_j(x|\theta^\mathcal{S},\theta^\mathcal{T},\mathcal{D})\Bigl),
    \end{aligned}
\end{equation}
for $i=2, 3, ..., k-1$. Although different pairs of blocks are trained separately, we can also put all these loss terms together as a summary:
\begin{equation} \label{eq:StableKD}
    \mathcal{L}_{StableKD} =\ell_{CE}+\lambda \ell_{KL}+\sum_{i=1}^{k-1} \ell_{MSE}^{\Hquad i}.
\end{equation}

Overall, Equation (\ref{eq:StableKD}) enables various kinds of knowledge transfer from the teacher blocks to their corresponding student blocks concurrently yet without interfering with each other. With only a few epochs, the loss shrinks quickly, and the students blocks start to converge.

\subsubsection{Recomposition.}\label{merge}
While the \emph{Decomposition} operation is designed to alleviate IBOE issue, the divided blocks are blind to each other. This leaves the student an architectural discrepancy between training and testing, and makes the student less accurate during testing, as we verified in Section~\ref{ab_study}. Therefore, we propose to progressively merge the divided blocks back along the training process. To be more concrete, we have the following definition:
\begin{definition}[Recomposing Function $\mathcal{R}$] Given $\mathcal{D}$ that encodes the information of 
$k$ blocks decomposed from a teacher or a student network,
$\mathcal{R}(\mathcal{D})$ merges the original $i$-th and $\left(i+1\right)$-th blocks together as a new block for every positive odd number $i\leq k-1$. 
\end{definition}
In other word, $\mathcal{R}(\cdot)$ is a function taking $\mathcal{D}$ as the input and  returning an updated $\mathcal{D}$, which has $\lceil \frac{k}{2}\rceil$ blocks after recomposition. Figure~\ref{fig:design}~(b) demonstrates this operation in which $i'=(i+1)/2$ for $i=1,3,..., k-1$, supposing $k-1$ is an odd number. Every time the recomposing function is performed, our StableKD enters a new stage of training, and finally evolves towards the vanilla KD.

\subsubsection{Complete pipeline.}\label{full}
\begin{algorithm}[tb]
\caption{StableKD algorithm}
\label{alg:algorithm}
\textbf{Parameter}: $\theta^\mathcal{S}, \theta^\mathcal{T}$\\
\textbf{Input}: decomposition variable $\mathcal{D}$, number of recompositions $n$, number of training epochs for each stage $e_i$, balancing factor of losses $\lambda$\\
\begin{algorithmic}[1]
\STATE Initialize $\theta^\mathcal{S}$ and freeze $\theta^\mathcal{T}$.
    \FOR{training stage $c=0, 1, ...,n$}
        \FOR{each of $e_c$ epochs}
            \STATE Update $\theta^\mathcal{S}$ with Equation (\ref{eq:StableKD}). \hfill\COMMENT{\textit{Decomposition.}}
        \ENDFOR
        \STATE $\mathcal{D}\gets \mathcal{R}(\mathcal{D})$ \hfill\COMMENT{\textit{Recomposition.}}
        \ENDFOR
        \STATE \textbf{return} $\theta^\mathcal{S}$
\end{algorithmic}
\end{algorithm}
Given the initial decomposition strategy $\mathcal{D}$ and the number of recompositions to perform $n$, we present the complete pipeline of StableKD in Algorithm~\ref{alg:algorithm}.
The complete pipeline is simple to understand and easy to implement.
It’s expected to achieve a higher accuracy and converge faster than vanilla KD. Moreover, 
our StableKD is as computation-efficient as the vanilla KD for each iteration, by avoiding complex and heavy data and architectural manipulations. Note that we denote \textbf{StableKD-k/n} as the full StableKD pipeline with $k$ divided blocks initially and with $n$ recompositions.

\section{Experimental Evaluation}\label{experiments}
In this section, we begin by validating 
the efficacy of StableKD in mitigating the IBOE issue. We then present empirical evidence showcasing the advantages of StableKD from three practical aspects: convergence speed, accuracy, and data efficiency. Our experimental analyses encompass convolutional neural networks and transformer-based models trained on diverse datasets, such as CIFAR100~\citep{krizhevsky2009learning}, Imagewoof~\citep{imagewang}, and ImageNet~\citep{russakovsky2015imagenet}. The results illustrate that StableKD achieves superior accuracy compared to previous approaches, while requiring fewer training epochs and reduced training data. Furthermore, we conduct ablation studies to assess the indispensability of the decomposition and recomposition operations in StableKD, and also explore potential avenues for future enhancements.

\subsection{StableKD in Alleviating IBOE}\label{stable}
\begin{figure}
    \centering
    \includegraphics[width=.85\linewidth]{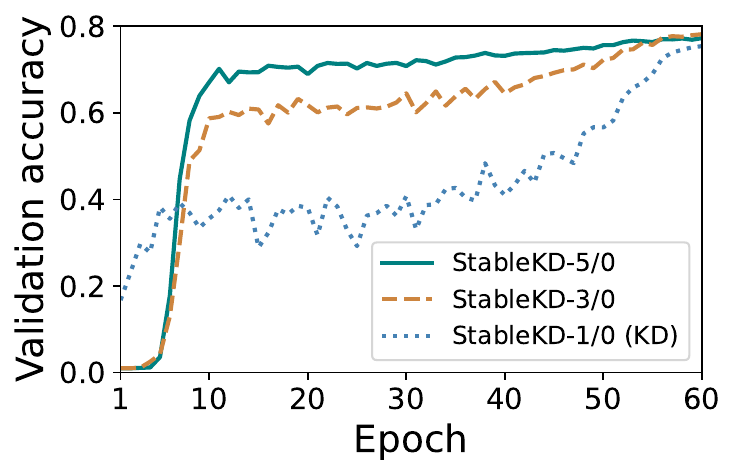}
    \caption{ 
    StableKD effectively alleviates the inter-block optimization entanglement.
    }
    \label{fig:blocks}
\end{figure}
This experiment aims to validate the efficacy of StableKD in mitigating the IBOE issue. We adopt the ResNet18~\citep{he2016deep} as the student and the ResNet34 with pre-trained accuracy $78.39\%$ as the teacher on CIFAR100. We explore three decomposition strategies with the values of k equal to 1, 3, and 5, while no recompositions are used. Note that setting $k = 1$ is equivalent to the vanilla KD. For all the three runs, we conduct the training for 60 epochs, with a learning rate of 0.05 and a OneCycleLR scheduler~\citep{smith2019super}. We monitor their validation accuracies along the training process. Figure~\ref{fig:blocks} shows that the accuracy of vanilla KD or StableKD-1/0 suffer from severe fluctuations, whereas the accuracy of StableKD-5/0 and StableKD-3/0 increase much more stably, effectively circumventing the IBOE issue.

\subsection{Illustrative Results}\label{main_result}
\begin{figure}
\centering
    \begin{subfigure}{\linewidth}
    \centering
    \includegraphics[width=.95\linewidth]{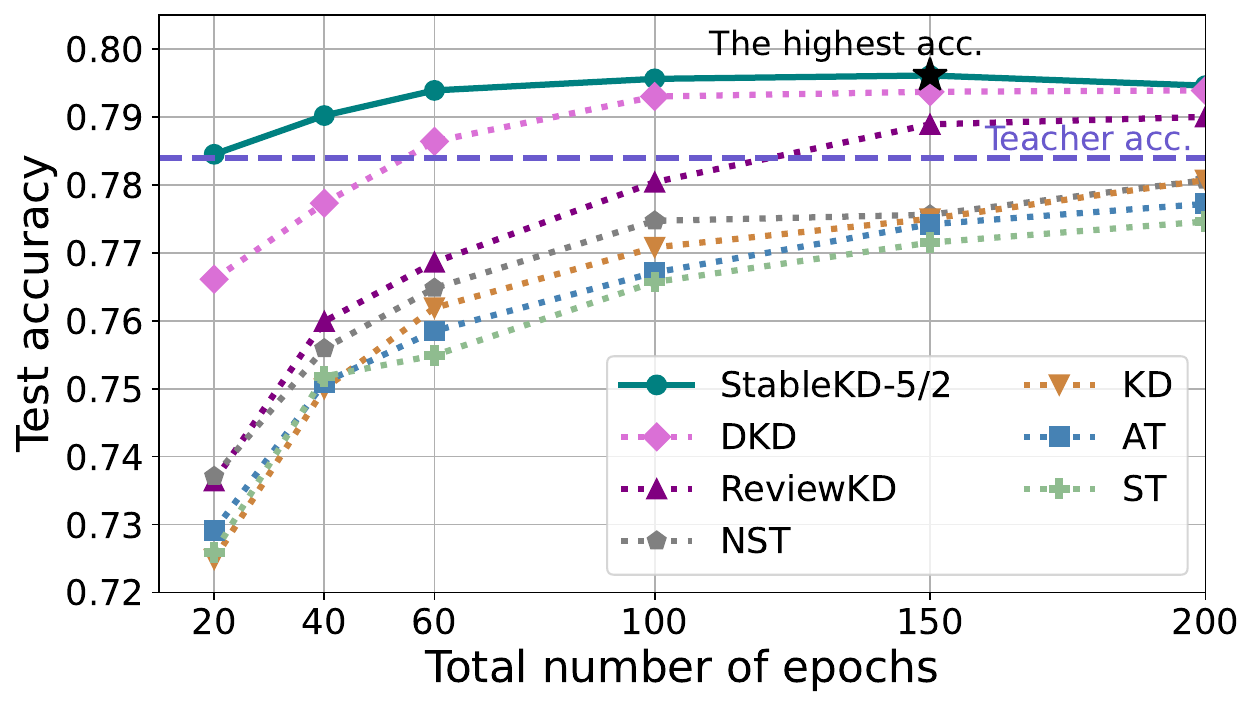}
        \caption{CIFAR100}
        \label{fig:cifar_resnet18}
    \end{subfigure}
    \begin{subfigure}{\linewidth}
        \centering
        \includegraphics[width=.95\linewidth]{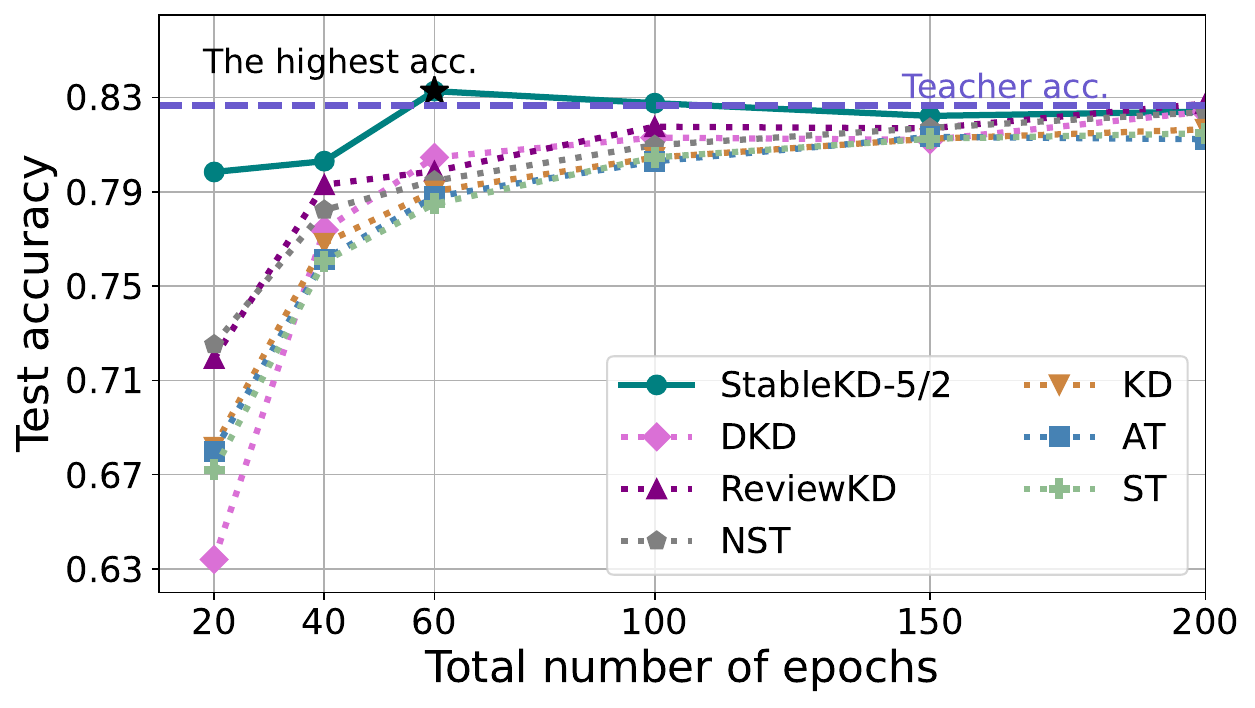}
        \caption{Imagewoof}
        \label{fig:IW_resnet18}
    \end{subfigure}
    \vspace{-0.25in}
    \caption{
    StableKD outperforms others by achieving higher test accuracy with fewer training epochs on CIFAR100 and Imagewoof datasets.
    }
    \label{fig:resnet18_acc}
    \vspace{10pt}
    \centering
    \begin{subfigure}{\linewidth}
        \centering\includegraphics[width=.95\linewidth]{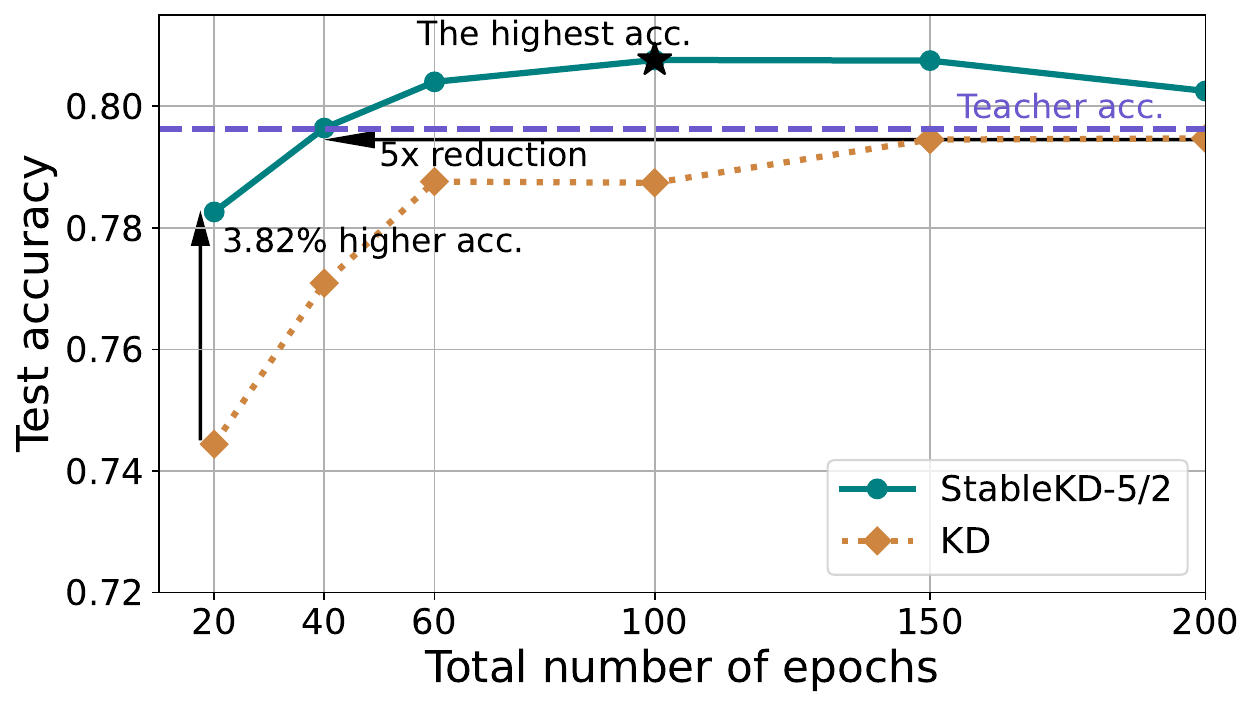}
        \caption{WRN16-10 \& WRN28-10}
        \label{fig:WRN}
    \end{subfigure}
    \begin{subfigure}{\linewidth}
        \centering\includegraphics[width=.95\linewidth]{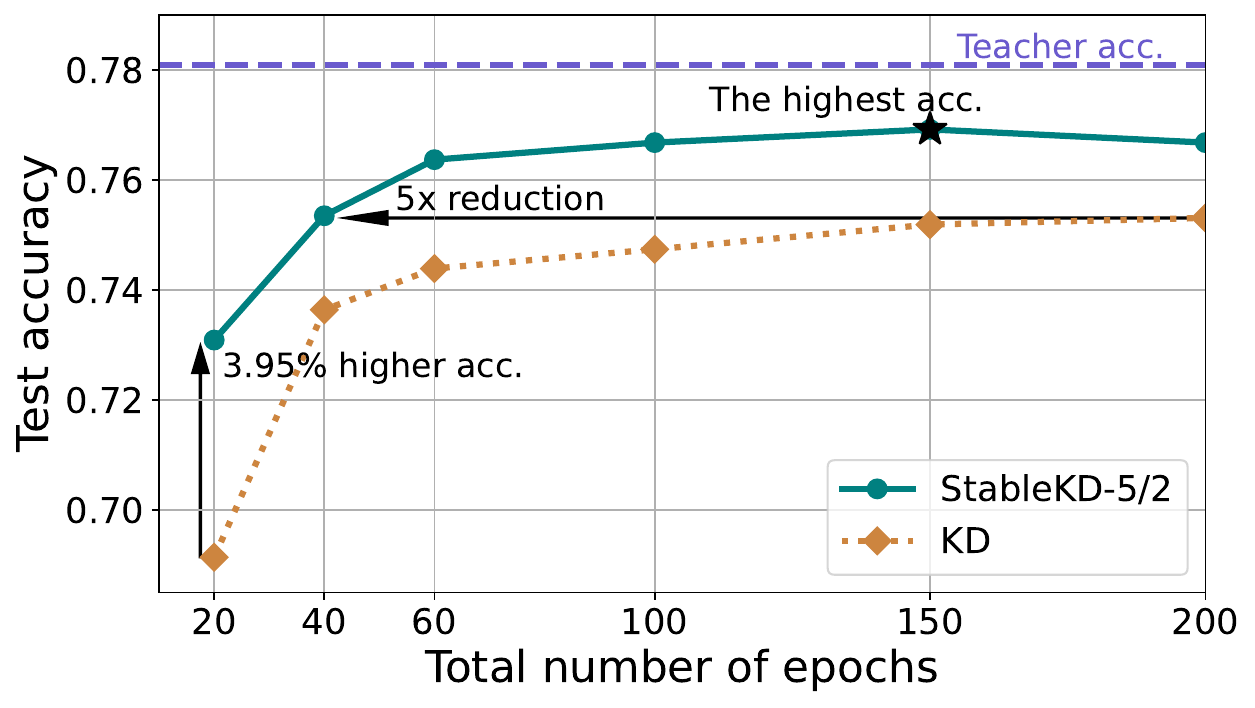}
        \caption{ResNeXt20($32\times 4d$) \& ResNeXt38($32\times 4d$)}
        \label{fig:resnext}
    \end{subfigure}
    \vspace{-0.25in}
    \caption{
    For different teacher-student pairs, StableKD consistently achieves higher accuracy on CIFAR100 with fewer epochs than the vanilla KD.
    }
    \label{fig:wrn_resnext}
\end{figure}

\subsubsection{Comparison of convergence speed and accuracy.}\label{convergence_acc}
In deep learning, the convergence speed generally refers to the number of epochs, iterations, or training time spent in achieving a high level of accuracy. By contrast, the accuracy is often reported given a fixed number of epochs. For comprehensive evaluation, here we mainly compare the trade-offs between the test accuracy and the total number of epochs. We first conduct experiments on CIFAR100 and Imagewoof datasets. We train a ResNet18 with a teacher ResNet34 for 20, 40, 60, 100, 150, and 200 epochs. For StableKD, we set the following hyper-parameters: $k=5, n=2, e_0=\lfloor e/3\rfloor, e_1=\lfloor e/3\rfloor, e_2=e-2\lfloor e/3\rfloor$, where $e$ is the total number of epochs. We compare StableKD against six representative works, including the vanilla KD~\citep{hinton2015distilling}, Attention Transfer (AT; \citealt{zagoruyko2016paying}), Sobolev Training (ST; \citealt{czarnecki2017sobolev}),  Neuron Selectivity Transfer (NST; \citealt{huang2017like}), Knowledge Review (ReviewKD; \citealt{chen2021distilling}), and Decoupled Knowledge Distillation (DKD; \citealt{zhao2022decoupled}).
Results in Figure~\ref{fig:resnet18_acc} show that StableKD renews the state of the art, consistently achieving the highest accuracy given varied number of epochs, where the gap is more evident with limited training budget. From another perspective, StableKD converges faster than others in achieving a high accuracy.

These conclusions are consistent when further applied to other teacher-student pairs, \emph{e.g.}, the WRN28-10-WRN16-10 and the ResNeXt38 ($32\times 4d$)-ResNeXt20 ($32\times 4d$). Results in Figure~\ref{fig:wrn_resnext} show that StableKD speeds up the KD process by 5$\times$, from 200 epochs to 40 epochs when trained on CIFAR100. With only 20 epochs, StableKD also obtains a significant 3.8\% and 4.0\% accuracy gain for the two pairs, respectively.

We also conduct experiments on ImageNet, a much larger scale of classification dataset with 1.2 million images. We use transformer-based backbones including ViT~\citep{dosovitskiy2020image} and Swin~\citep{liu2021swin}. Results are summarized in Table~\ref{tab:in_90}. Compared to training w/o KD, vanilla KD, and DeiT~\citep{touvron2021training}, StableKD improves the accuracy by 2.0\% - 3.7\% for both 90 and 300 epochs. StableKD-5/3 trained with 90 epochs even beats the vanilla KD method with 300 epochs (81.51\% \emph{v.s.} 81.15\%).

\begin{table}
\centering
\resizebox{.99\linewidth}{!}{%
\begin{tabular}{c|c|c|c|c}
\toprule
    Student $\mathcal{S}$ &Teacher $\mathcal{T}$& Epochs & KD Scheme & ${\text{Acc.}}$ (\%) \\ \midrule
    \multirow{4}*{ViT-S/8} & \multirow{4}*{\begin{tabular}{@{}c@{}}ViT-S/16 \\ (74.03\%)\end{tabular}}& \multirow{4}*{90} &  w/o KD   & 74.00 \\
    & & &KD   & 75.05     \\
    & & &DeiT & 75.45     \\
    & & &StableKD-5/3  & \textbf{77.72} \\ \cmidrule(lr){1-5}
    \multirow{3}*{Swin-T} &\multirow{3}*{\begin{tabular}{@{}c@{}}Swin-S \\ (83.33\%)\end{tabular}}& \multirow{3}*{90} & w/o KD    & 79.53   \\
    && &KD  & 79.95    \\
    && &StableKD-5/3 & \textbf{81.51}  \\ \cmidrule(lr){1-5}
    \multirow{3}*{Swin-T} &\multirow{3}*{\begin{tabular}{@{}c@{}}Swin-S \\ (83.33\%)\end{tabular}}& \multirow{3}*{300} & w/o KD    & 80.37   \\
    && &KD  &  81.15    \\
    && &StableKD-5/3 & \textbf{82.61}  \\ \bottomrule
\end{tabular}
}
\caption{Performance comparison on the large scale ImageNet dataset. ViT-S/8 denotes a network with depth 8 derived from ViT-S/16. DeiT is designed for ViT and not directly applicable to Swin.
}
    \label{tab:in_90}
\end{table}

\subsubsection{Comparison of data efficiency.}
\begin{table*}
    \centering
    \small
    \resizebox{.88\linewidth}{!}{%
    \begin{tabular}{c|cc|c|c|ccccc}\toprule
    \multirow{2}*{Dataset}&\multirow{2}*{Student $\mathcal{S}$}& \multirow{2}*{Teacher $\mathcal{T}$}&\multirow{2}*{\# epochs}&\multirow{2}*{\begin{tabular}{@{}c@{}}KD \\ Scheme\end{tabular}} & \multicolumn{5}{c}{Acc. (\%) for various subset sizes}\\ 
    &&&&   & $20\%$  & $40\%$ & $60\%$  & $80\%$  & 100\% \\\midrule
    
     \multirow{7}*{CIFAR100}&\multirow{7}*{ResNet18}& \multirow{7}*{\begin{tabular}{@{}c@{}}ResNet34 \\ (78.39\%)\end{tabular}}&\multirow{7}*{300}&KD   & 55.65 & 67.64 & 72.88 & 75.91 & 78.12 \\
     &&&&AT& 56.53 & 67.22 & 71.99 & 75.31 & 77.99 \\
     &&&&ST& 55.47 & 67.54 & 71.87 & 75.26 & 77.75\\
     &&&&NST  & 60.84 & 69.87 & 74.36 & 76.16 & 78.14 \\
     &&&&ReviewKD & 62.20 & 71.25 & 75.20 & 77.19& 78.97\\
     &&&&DKD  & 69.76 & 75.60 & 77.39 & 78.77 & 79.39 \\
     &&&&StableKD-5/2  & \textbf{76.10} & \textbf{78.50} & \textbf{79.18} & \textbf{79.63} & \textbf{79.41} \\ \midrule
     \multirow{3}*{ImageNet}&\multirow{3}*{Swin-T}& \multirow{3}*{\begin{tabular}{@{}c@{}}Swin-S \\ (83.33\%)\end{tabular}}& \multirow{3}*{90}&w/o KD   & 66.01 & 74.98 & 77.76 & 78.67 & 79.53 \\
     &&&&KD  & 69.64 & 76.10 & 78.33 & 79.52 & 79.95 \\
     &&&&StableKD-5/3  & \textbf{76.03} & \textbf{78.87} & \textbf{80.34} & \textbf{80.89} & \textbf{81.51} \\ 
     \bottomrule
    \end{tabular}
    }
    \caption{Comparison of data efficiency with varying subsets of the training data.}
    \label{tab:DataEff}
\end{table*}     
We compare the data efficiency by examining how well different KD methods perform with reduced training dataset, from 100\% to only 20\% of the whole dataset. Other training settings and baselines are the same to previous experiments. We present the results in Table~\ref{tab:DataEff}. Compared to prior works, StableKD demonstrates significantly higher data efficiency by maintaining robustly good accuracy with varying subset sizes. With 20\% of the training data, StableKD outperforms others by a large margin ranging from 6\% to 20\% in accuracy. Furthermore, StableKD trained with 40\% - 60\% of the training data performs similarly or even better than others trained with the whole dataset.

It is common practice to pre-train transformer-based models on external data before fine-tuning them on the target dataset to prevent overfitting and improve performance. To further validate the strengths of StableKD in achieving higher data efficiency and thereby reducing the training cost, we trained transformer-based student networks from scratch on CIFAR100 without any pre-training on external data. Results in Table~\ref{tab:in2cifar} show that all baselines perform poorly while StableKD stands out with 10\% - 18\% significantly higher accuracy, indicating its superior data efficiency.

\subsection{Ablation Studies}\label{ab_study}
\paragraph{Effect of decomposition.}
To validate the effectiveness of decomposition, we degrade StableKD to conventional feature-based KD methods by imposing the feature alignment but removing the independence of different blocks. We adopt the ResNet18-ResNet34 pair and train the student for 60 epochs. Results in Table~\ref{tab:ab_div} show that the feature alignment through MSE loss does enhance the vanilla KD by 1.0\% in accuracy. However, a more substantial improvement of 3.2\% is achieved with the implementation of StableKD. This, therefore, substantiates the necessity of our decomposition operation, particularly the importance of enabling \emph{independent} block-wise KD.

\begin{table}
\centering
\footnotesize
\resizebox{.9\linewidth}{!}{%
\begin{tabular}{c|c|c|c}
\toprule
    Student $\mathcal{S}$& Teacher $\mathcal{T}$ & Scheme & ${\text{Acc.}}$ (\%) \\ \midrule
    \multirow{4}*{ViT-T/8} & \multirow{4}*{\begin{tabular}{@{}c@{}}ViT-T/16 \\ (88.98\%)\end{tabular}}&  w/o KD   & 67.99 \\
    & &KD   & 66.46     \\
    & &DeiT & 71.01     \\
    & &StableKD-5/2  & \textbf{84.18} \\ \cmidrule(lr){1-4}
    \multirow{3}*{Swin-T} &\multirow{3}*{\begin{tabular}{@{}c@{}}Swin-S \\ (90.39\%)\end{tabular}}& w/o KD    & 73.78   \\
    &&KD  & 73.84    \\
    &&StableKD-5/2 & \textbf{83.92}  \\ \bottomrule
\end{tabular}
}
\caption{Comparison when training transformer-based networks on CIFAR100 from scratch without pre-training on external data. We use 600 training epochs here.}
    \label{tab:in2cifar}
\end{table}

\begin{table}[t]
\centering
\footnotesize
\resizebox{.85\linewidth}{!}{%
\begin{tabular}{c|ccc}
\toprule
    Scheme &KD  & KD + MSE & StableKD-5/2\\ \midrule
    ${\text{Acc.}}$ (\%)&76.19&77.15&\textbf{79.39}\\
       \bottomrule
\end{tabular}
}
\caption{With feature alignment, KD + MSE is still inferior to StableKD.}
\label{tab:ab_div}
\end{table}

\paragraph{Effect of recomposition.}
We next evaluate the effectiveness of progressively merging the decomposed blocks by training w/ and w/o recomposition. Other training settings remain the same with the previous ablation experiment.
Results in Table~\ref{tab:ablation_block} show that, without recomposition, StableKD-5/0 surpasses the vanilla KD for 20, 40, and 60 epochs due the use of decomposition in accelerating the convergence. However, StableKD-5/0 gets lower accuracy than the vanilla KD for more training epochs. This is because individual blocks are trained independently but lack mutual awareness, leading to sub-optimal accuracy. To address it, we introduce the recomposition procedure, which consistently yields higher accuracy as is evident in the results.

\begin{table}[t]
    \centering
    \small
    \resizebox{1.\linewidth}{!}{%
    \begin{tabularx}{\linewidth}{
p{\dimexpr.28\linewidth-2\tabcolsep-1.3333\arrayrulewidth}
p{\dimexpr.12\linewidth-2\tabcolsep-1.3333\arrayrulewidth}
p{\dimexpr.12\linewidth-2\tabcolsep-1.3333\arrayrulewidth}
p{\dimexpr.12\linewidth-2\tabcolsep-1.3333\arrayrulewidth}
p{\dimexpr.12\linewidth-2\tabcolsep-1.3333\arrayrulewidth}
p{\dimexpr.12\linewidth-2\tabcolsep-1.3333\arrayrulewidth}
p{\dimexpr.12\linewidth-2\tabcolsep-1.3333\arrayrulewidth}}
    \toprule
    \centering KD & \multicolumn{6}{c}{Acc. (\%) for various numbers of epochs} \\
    \cmidrule(lr){2-7}
    \centering schemes   & \centering20  & \centering40 & \centering60  &\centering 100  & \centering150 & \centering\arraybackslash 200 \\\midrule
    
     \centering KD   & \centering 72.49 & \centering 75.00 &\centering 76.19 &\centering 77.08 &\centering\arraybackslash 77.50 & 78.07 \\
     \centering StableKD-5/0 &\centering 76.59 &\centering 76.90 & \centering 77.26 &\centering 76.99 & \centering 77.39 &\centering\arraybackslash 77.23 \\
     \centering StableKD-5/2  &\centering \textbf{78.45} &\centering \textbf{79.02} &\centering \textbf{79.39} &\centering \textbf{79.21} & \centering \textbf{79.61} &\centering\arraybackslash \textbf{79.46} \\
     \bottomrule
    \end{tabularx}
    }
    \caption{StableKD-5/2 with recomposition consistently obtains higher accuracy than StableKD-5/0 without recomposition and the vanilla KD.}
    \label{tab:ablation_block}
\end{table}

\begin{table}
\centering
\footnotesize
\resizebox{\linewidth}{!}{%
\begin{tabular}{cccccc}
\toprule
    \multirow{2}*{\large{Student}} & \large{Teacher} & \large{Projector} & \multicolumn{3}{c}{\large{Test accuracy (\%)}}\\\cmidrule(lr){4-6}
    &(Acc., \# params)&(\# params)&w/o KD& KD & StableKD-2/1\\
    \midrule
        \multirow{2}*{\large{WRN40-2}}& \large{WRN40-4} &\XSolidBrush (2.2M)&\large{74.56}&\large{75.35}& \large{n/a}\\
        &($77.11\%$, 8.9M)& \CheckmarkBold (2.3M)&\large{74.96}&\large{75.40}&\large{\textbf{75.61}}\\\midrule
        \multirow{2}*{\large{WRN16-8}} & \large{WRN16-10} & \XSolidBrush (11.0M)& \large{78.53}&\large{78.67}& \large{n/a}\\
        &($79.69\%$, 17.1M)&\CheckmarkBold (11.6M)&\large{78.75}&\large{78.73}&\large{\textbf{79.57}}\\
       \bottomrule
\end{tabular}
}
\caption{More results for heterogeneous architectures. The reported accuracies of WRN40-2 and WRN16-8 in their original paper are $73.96\%$ and $77.93\%$, respectively.}
    \label{tab:mix}
\end{table}
\paragraph{Dealing with heterogeneous feature dimensions.}
Our main experiments are conducted when the student and teacher networks have the same feature dimension in each block. What if the dimension differs, \emph{e.g}, having different network width?  To address this and extend the application scope of StableKD, we can simply insert additional projectors, \emph{e.g.}, the conv $1\times 1$ layer, within the student network, and thus enable the use of our StableKD. To validate this, we conduct experiments on Wide ResNet~\citep{zagoruyko2016wide} with varying width factors. We train them on CIFAR100 for 200 epochs. Results in Table~\ref{tab:mix} demonstrate the promising applicability of StableKD to heterogeneous feature dimensions. 

\section{Conclusion}
In this paper, we have studied and uncovered a common issue of existing KD approaches -- the inter-block optimization entanglement that causes their slow convergence and lower accuracy. To overcome it, we have proposed a novel StableKD framework, which divides a pair of teacher and student networks into several blocks, and distills the block-wise knowledge separately before combining them together for inference. We have conducted extensive experiments and validated that StableKD is highly accurate, fast-converging, and data-efficient. We hope that our StableKD can make KD more practical as a tool for model compression and acceleration, particularly in the era of building foundation models towards artificial general intelligence.
\appendix
\vspace{3mm}
\begin{center}
\noindent{\LARGE\textbf{Appendix}}
\end{center}

\section{Limitations and Future Work}\label{limit}
We have verified the application of StableKD in image classification across several different neural networks. However, the proposed decomposition can become complex when the model architecture includes longer shortcuts, \emph{i.e.}, operations that pass features from shallow layers to deep ones. For example, networks for semantic segmentation, such as FCN \citep{long2015fully} and U-Net \citep{ronneberger2015u}, often concatenate features from shallow and deep layers to enhance performance. Since our proposed StableKD primarily focuses on dividing blocks in a sequential order, the exploration of non-sequential neural networks remains an open area for future work.

\section{Training Details}\label{training}
\begin{figure}
    \centering
    \includegraphics[width=.8\linewidth]{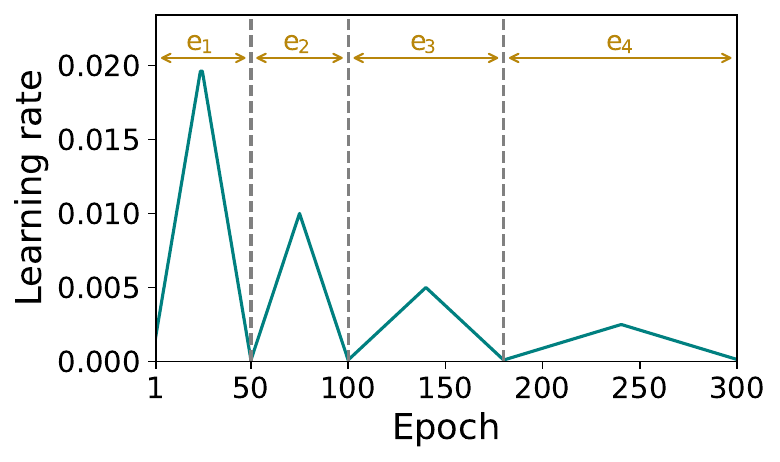}
    \vspace{-0.1in}
    \caption{Learning rate scheduler used in StableKD.}
    \label{fig:scheduler}
\end{figure}

\paragraph{Learning rate scheduler.} We propose a novel scheduler to enable per-training-stage warm-up and annealing. Intuitively, our learning rate extends the concepts from the cyclical learning rate~\citep{smith2017cyclical}  in PyTorch CyclicLR with \textit{triangular2} strategy; that is, instead of using cycles with same number of epochs, each cycle corresponds to the number of epochs in each StableKD stage. Illustration can be found in Figure~\ref{fig:scheduler}. For other comparison schemes, we use a OneCycle scheduler~\citep{smith2019super} in the CIFAR100 and Imagewoof experiments (Figure~\ref{fig:resnet18_acc}, \ref{fig:wrn_resnext}, Table~\ref{tab:DataEff}, \ref{tab:in2cifar}, \ref{tab:ab_div}, \ref{tab:ablation_block}, \ref{tab:mix}), and a cosine-annealing scheduler with linear warm-up in the ImageNet experiments (Table~\ref{tab:in_90},\ref{tab:DataEff}). Note that all experiments in Figure~\ref{fig:blocks} are conducted with a OneCycle scheduler.

\paragraph{Optimizer.} We adopt SGD as the optimizer in the CIFAR100 and Imagewoof experiments (Figure~\ref{fig:blocks}, \ref{fig:resnet18_acc}, \ref{fig:wrn_resnext}, Table~\ref{tab:DataEff}, \ref{tab:in2cifar}, \ref{tab:ab_div}, \ref{tab:ablation_block}, \ref{tab:mix}), and use AdamW for the ImageNet experiments (Table~\ref{tab:in_90}, \ref{tab:DataEff}).

\paragraph{Hyperparameters.} Our empirical experience recommends a higher learning rate for StableKD. In the CIFAR100 and Imagewoof experiments, all the experiments are conducted with a batch size of $128$ and weight decay of $5e-4$. The maximal learning rate of StableKD is set to $0.5$. For other comparison schemes, we use a maximal learning rate of $0.1$, where they demonstrate better performance and thus serve as stronger baselines. In the ImageNet experiments, we set the maximal learning rate, batch size, and weight decay to $2e-2, 256, 2e-1$ for StableKD, and to $2e-3, 512, 4e-2$ for other schemes, respectively. We found that StableKD generally performs better when a larger learning rate, a smaller batch size, and a smaller weight decay are used. This differs from previous works, which a smaller batch size usually accompanies a larger weight decay and a smaller learning rate. We attribute this difference to the benefits of higher stability that make the model easier to train.

\begin{figure}[h]
    \centering
    \includegraphics[width=\linewidth]{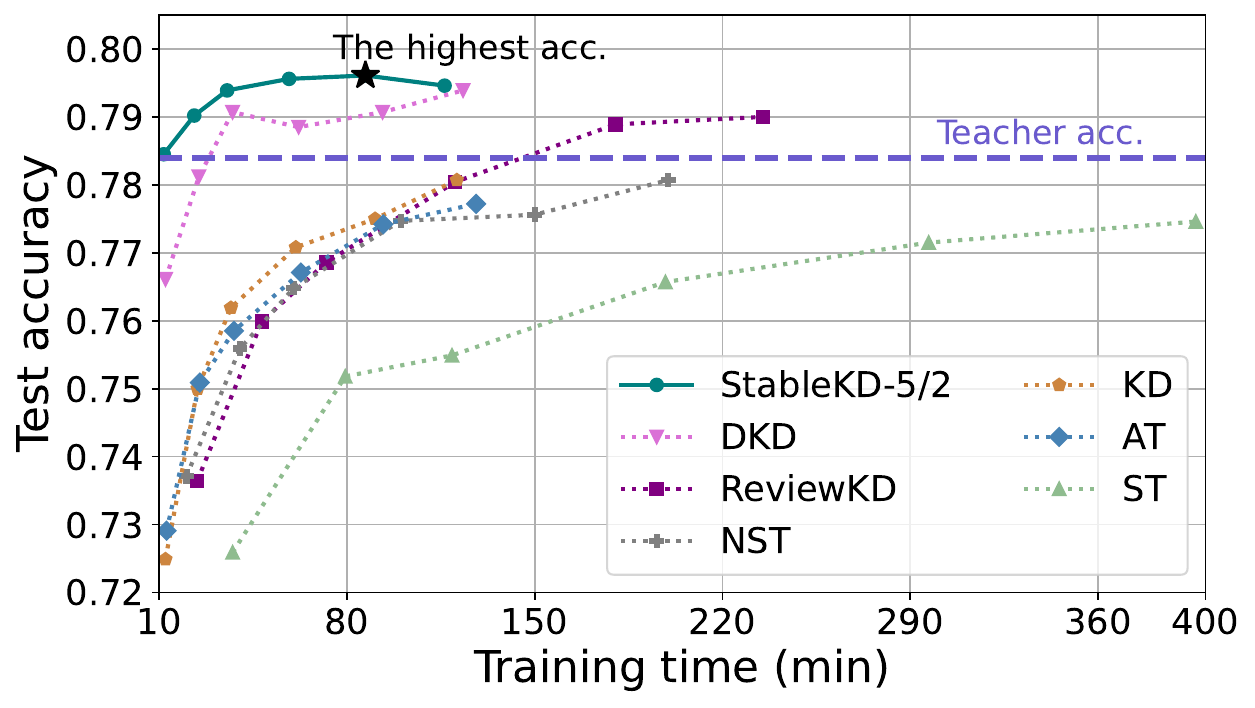}
    \vspace{-0.25in}
    \caption{Comparison of accuracy on CIFAR100 with the ResNet34-ResNet18 pair.}
    \label{fig:time_acc}
\end{figure}
\begin{table}[!h]
    \centering
    \begin{tabular}{ccc}
            \toprule
            KD  & \multicolumn{2}{c}{Average time per epoch (s)}\\
            \cmidrule(lr){2-3}
            schemes & CIFAR100&Imagewoof\\
            \cmidrule(lr){1-3}
            KD     & 36.3   & 14.9   \\
            AT     & 37.5   & 15.6  \\
            ST     & 120.6  & 44.7   \\
            NST    & 60.0   & 30.0\\
            ReviewKD & 72.0 & 28.4\\
            DKD & 37.3 & 15.3\\
            StableKD-5/2    & \textbf{35.4}   & \textbf{14.7}   \\
            \bottomrule
    \end{tabular}
    \vspace{-0.05in}
    \caption{Comparison of time cost per epoch with the ResNet34-ResNet18 pair.}
    \label{tab:runtime}
\end{table}

\section{Time Cost}\label{time_cost}
StableKD has low computational complexity and is easy to implement, as no overlapping backward process takes place in the student network. Table~\ref{tab:runtime} shows the average training time per epoch on the CIFAR100 and Imagewoof datasets, with a teacher ResNet34 and a student ResNet18 on a single RTX-2080Ti GPU. With these results, we can also adjust the horizontal axis of Figure~\ref{fig:resnet18_acc} to display time cost for clearer visualization. For example, Figure~\ref{fig:time_acc} shows the results on CIFAR100 with the same teacher-student pair.

\section{Decomposition Details}\label{decomp}
This paragraph describes the decomposition recipes of StabelKD in the main paper. In our experiment, we regard the last MLP layer as a single block. For ResNet~\citep{he2016deep}, Wide ResNet~\citep{zagoruyko2016wide}, and ResNeXt~\citep{xie2017aggregated} models, we divide the convolutional layers based on the dimension of each residual block. For example, the first decomposed block of ResNet18 in StableKD-5/n consists of the layers from first convolution layer to the last one with an output channel of 64. The second, third, and fourth blocks
 include the layers with output channelS of 128, 256, and 512, respectively. The fifth block represents the fully-connected layer. A similar recipe is used in ResNet34 as well. For Swin models, we follow the division method in its original paper~\citep{liu2021swin}, where the attention layers are decomposed into blocks with depths of 2, 2, 6, 2 in Swin-T, and 2, 2, 18, 2 in Swin-S. For ViT models~\citep{dosovitskiy2020image}, the attention layers are equally distributed to the first four blocks of StableKD-5/n, and the last block contains the fully-connected layer.

\bibliographystyle{named}
\bibliography{ijcai24}

\end{document}